# Continuous Monitoring for Road Flooding With Satellite Onboard Computing For Navigation for OrbitalAI Φsat-2 challenge

Vishesh Vatsal, Gouranga Nandi and Primo Manilal, *Member, SkyServe*

***Abstract*—Continuous monitoring for road flooding could be achieved through onboard computing of satellite imagery to generate near real-time insights made available to generate dynamic information for maps used for navigation. Given the existing computing hardware like the one considered for the PhiSat-2 mission, the paper describes the feasibility of running the road flooding detection. The simulated onboard imagery dataset development and its annotation process for the OrbitalAI Φsat-2 challenge is described. The flooding events in the city of Bengaluru, India were considered for this challenge. This is followed by the model architecture selection, training, optimization and accuracy results for the model. The results indicate that it is possible to build low size, high accuracy models for the road flooding use case.**

## I. INTRODUCTION

Commonly used navigation tools (like Google Maps, Waze, Apple Maps etc.) for route planning use road network generated from dated satellite-derived and survey maps. The satellite-derived high resolution imagery on a navigation basemap could be on average 6 months to 5 years depending on the area [1]. Dated maps used for navigation often miss the following dynamic road health parameters: flooded road segments, recent road damage, snow cover or mud cover (or landslide), blockages due to repairs among other issues. This often leads to consequences of traffic congestion, accidents, delay in the Average Transit Time for general people, businesses and even emergency medical services. An average driver loses 29 hours a year [2] due to bad maps, which means that just for the United States, a loss of billions of US dollars. Globally, the above number maybe a much, much larger value. Earth Observation is well-poised to contribute a solution to this problem.

The lowering cost of launch is helping Earth Observation industry to launch satellites at a rate which can get them to build constellations with high resolution with target revisit rate between 5 minutes and 1 hour [3][4][5]. With the capability of onboard computing performed on the latest satellite imagery [6][7][8], it is also expected that onboard processing will significantly reduce bandwidth costs of communication between earth observation satellites and ground-stations due to compressed information being downlinked [9]. This promises enhanced efficiency, agility, autonomy, and reconfigurability in Earth Observation. End users and applications also require valuable insights and optimal decisions with minimal delay. Hence, significant research efforts have been dedicated to exploring onboard intelligence for Earth Observation applications, such as early detection of natural disasters, vessel incidents, and gas leaks.

Onboard processing in satellites, as a response, is revolutionising space data by enabling faster and more efficient data transmission, reducing the dependence on ground-based processing, and enabling real-time decision-making. Onboard intelligence empowers us to identify low-quality Earth Observation data, such as cloud-covered satellite images or remote sensing images with limited relevant information and discard them. This not only saves costs but also minimizes the need for data transmission to Earth. High revisit rates and lowered bandwidth costs open the domain of vehicle navigation to earth observation. This study proposes a robust, optimized water segmentation model to continuously downlink road health data for generating dynamic maps used for navigation.

## II. CONCEPT FOR ROAD HEALTH MONITORING FROM EARTH OBSERVATION

The concept proposed in the paper relies on point to point navigation based on existing Application Programming Interfaces (APIs) which ingests source location from the current location predicted by the Global Navigation Satellite System (GNSS) system on a moving vehicle as well as the destination to go and provides a route based on existing basemaps. This is shown as point 1 in Figure 1 below. Lets consider a scenario where a road segment lying on the predicted route is flooded and as shown as point 2. Following this, a CubeSat from a satellite constellation with onboard computing capabilities acquires a multispectral image of the entire city.

A road health classification model (in this case, a water segmentation model) identifies the flooded segment as a water mask and downlinks the georeferenced mask to the ground as illustrated in Point 3 in Figure 1. The geo-referenced locations of flooded road segments which are intersected with road networks from existing basemaps. This timestamped intersection location information is updated in the road network graph weights used by the navigation API service provider. Based on this, the navigation API updates an optimal route plan to the destination avoiding the flooded road segment as shown in Points 5 and 6 in Figure 1. If implemented in multiple units of a multispectral satellite constellation, a generalized road flood detection can provide revisits in the order of minutes. This can lead to generation of dynamic basemaps over a global scale which can eliminate at least some causes of bad route plans.

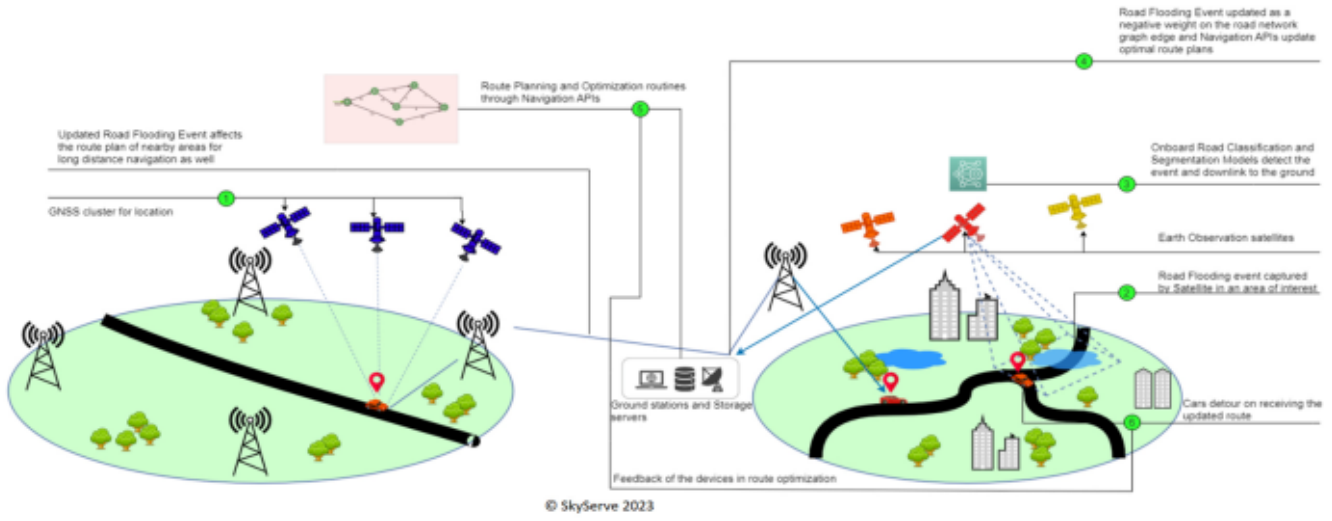

**Fig.1.** Illustrative schematic for a road flood detection application for a navigation use case.

To simulate its possibility, a test was done for a constellation of satellites flying over existing global road network. Under the Global Roads Inventory Project, Meijer et.al. [10] provide a global raster with each cell representing a 5 x 5 arcminute resolution (approximately 8 km x 8 km at the equator) mapped to road density as road length per unit area (m per sq.km.). Highways, primary roads, secondary roads, tertiary roads and local roads and unspecified roads are all given equal weights to produce this density map. This density is visualized in Figure 2.

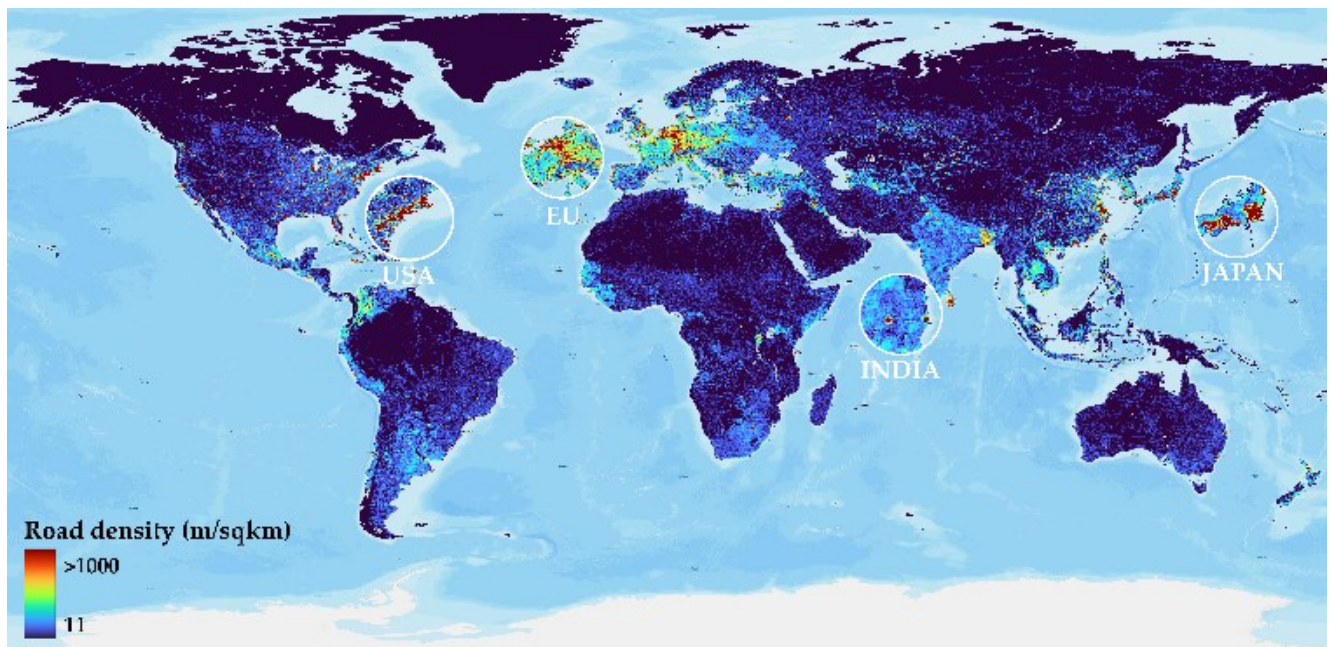

**Fig.2.** Global Road Inventory Project Road density raster.

Two batches of Planet Labs' Dove [11] satellites launched on 13th January 2022 and 3rd January 2023 in Sun Synchronous Orbit are considered for study. Cumulatively, the two batches comprise 77 satellites. Each satellite has a swath of 8 km or better which gives the satellite an ability to cover each cell (8 km x 8 km) in a single image. Also, satellites in Lower Earth Orbit (LEO) fly around 7.8 km/s at 500 km altitude and thereby can cover each cell of 8 km x 8 km in approximately a second.

Assume each satellite has onboard processing capability which can generate the near-realtime flood In this simulation, the ground track for all the 77 satellites is propagated at 1 sec time step between 1st February and 2nd March 2024 using the Two-Line Elements lying within this time-range. The revisit count for each 8 km x 8 km cell is initially set to zero. At each time step of the propagation for an individual satellite, if the sub-satellite point is found lying within a cell having road density greater than 10 m per sq.km. and the local time is between 8 AM and 5 PM (daylight view), the revisit count for that cell is incremented by 1. This is repeated for all satellites to get a total constellation revisit count raster image shown in Figure 3.

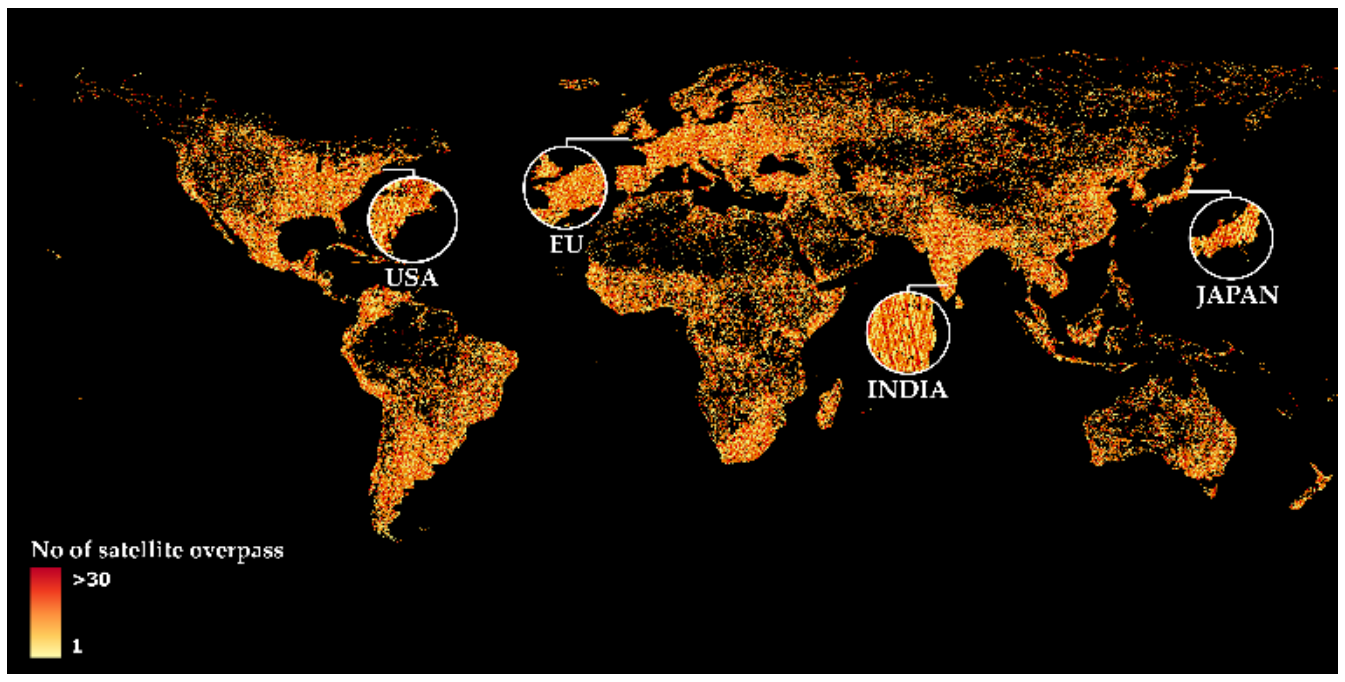

**Fig.3.** Daytime revisit Map for the constellation over every road cell with density greater than 10 m per sq.km.

This simulation suggests that global roads are covered well by this constellation with many areas exceeding 30 revisits per month. As seen in Figure 4, some revisit counts reach around 40 per month. This translates to more than 1 revisit per day.

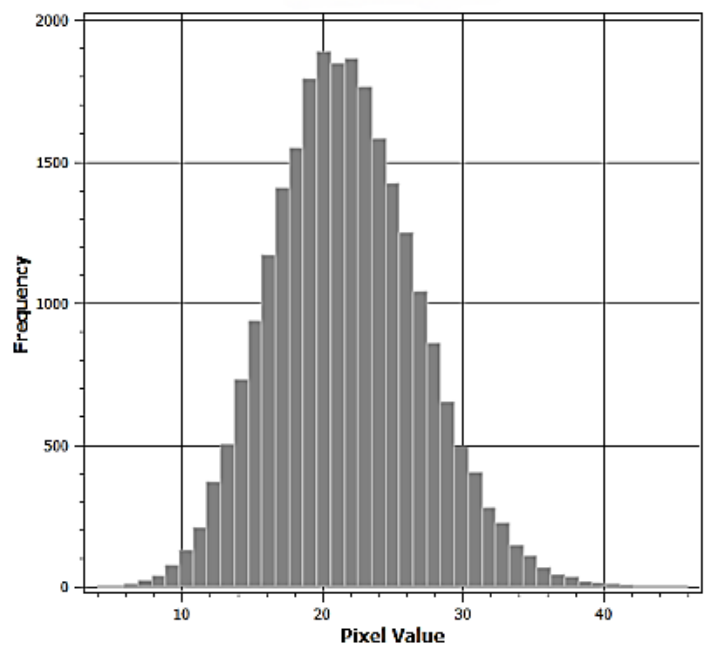

**Fig.4.** Histogram of revisit count for the constellation over all road cells having a minimum of 1 revisit and density greater than 10 m per sq.km.

Over the month of study period, the mean cell revisit count is 21.65 and standard deviation of 5.32 revisits. This suggests that broadly repurposing a subset of an existing satellite constellation with onboard processing can provide a global, daily road-flooding detection capability on average. With constellations of thousands of satellites, this capability would lead to higher revisits (minutes to hours).

## II. ORBITAL AI PHISAT-2 CHALLENGE

The European Space Agency (ESA), with its vision of creating a thriving ecosystem of Earth observation applications using edge computing in space, had launched the OrbitalAI Φsat-2 challenge [12] as a global competition. This challenge addressed the existing obstacles in the flow of Earth observation data and paves the way for the next generation of applications.

The challenge began with the first phase between 15th February to 20th September 2023. For this competition, the participants will use the challenge simulated dataset and create a working prototype model adapted to run onboard Φsat-2. The challenge provided the participants with a Φsat-2 simulator [13] to generate representative L1C onboard imagery from open-access Sentinel-2 L1C imagery. The L1C imagery referred to Top of Atmosphere Reflectance in sensor geometry, fine geo-referenced, fine band-to-band alignment (<10 m RMSE). As opposed to Sentinel-2 nomenclature, this product level was not orthorectified. SkyServe proposed the Road Health Monitoring Model for flooded road segment identification in the challenge.

### *A. PhiSat-2 simulator*

The PhiSat-2 simulator's essential steps relevant for the paper are described in Figure 1. In the interest of simulating any area of interest from nadir-facing geometry and not have any cost constraints [13], Sentinel-2 Multispectral sensor's L1C reflectance product is chosen as an input to the simulator [14].

The simulator's workflow is described in Figure 5. The simulator takes the 10m-Ground Sampling Distance (GSD) and implements a bicubic interpolation to get to the representative 4.75m-GSD image output for all the eight bands. The next step involves a band misalignment. Due to the pushbroom acquisition mode [13], the various bands over a given area are not acquired at the time.

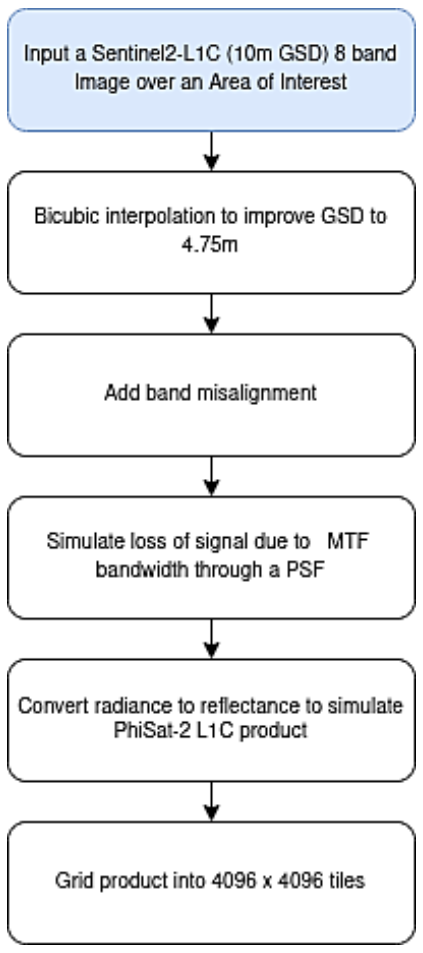

**Fig.5.** PhiSat-2 simulator workflow

As such, they suffer from platform attitude stability during the various acquisition times (the misalignment due to velocity is compensated onboard by the payload). Keeping a specific band as reference, a misalignment direction and magnitude is chosen as random number extracted from a normal distribution to shift each of the bands. To account for the system Modulation Transfer Function (payload + platform), the next step consists of the convolution of the input image with a filter function representing the system specific Point Spread Function (PSF). Basis the solar irradiance, Sun-zenith angle and Earth-Sun distance, the multiplicative factor needed to convert radiance to reflectance is computed to simulate the L1C product. The last step involves tiling the image into 4096 x4096-sized grids as the final product available for processing onboard..

## III. DATASET PREPARATION

The images initially extracted from the simulator were for four bands: Red, Green, Blue, and NIR (Near Infrared) with their corresponding masks. The images, in false colour, look like the one sample shown in Figure 6. The normalized difference water index (NDWI) [15] is used to identify water content in water bodies. The resulting NDWI is thresholded by the value >0.01. Subsequently, the identified areas are categorized into water and non-water. To refine initial detection inaccuracies, a manual correction process is implemented through visual interpretation, thereby enhancing the accuracy of the machine learning (ML) model. The dataset, comprising water bodies observed on three distinct dates for Bengaluru city (3rd May, 5th September, and 25th October 2022), is then processed to accurately segment surface water bodies into 256 x 256 chips, finally creating a training dataset for the ML model. The NDWI is computed using a specified equation (1):

$$NDWI = \frac{GREEN-NIR}{GREEN+NIR}. \qquad (1)$$

Where, GREEN and NIR is band 3 and 4 respectively. Each image for the dataset was finally chosen to be of 4 channels: RED, GREEN, BLUE, NDWI.

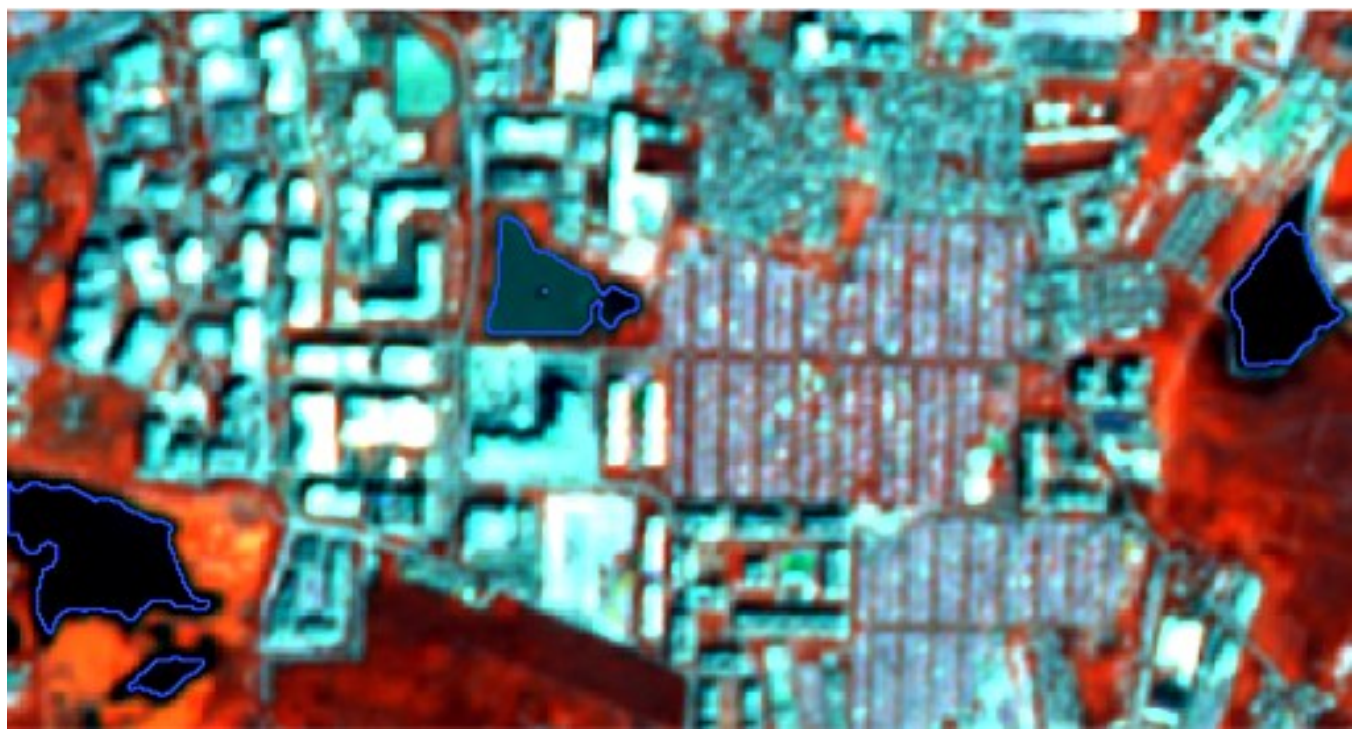

**Fig.6.** PhiSat-2 simulator sample image in false colored composite view

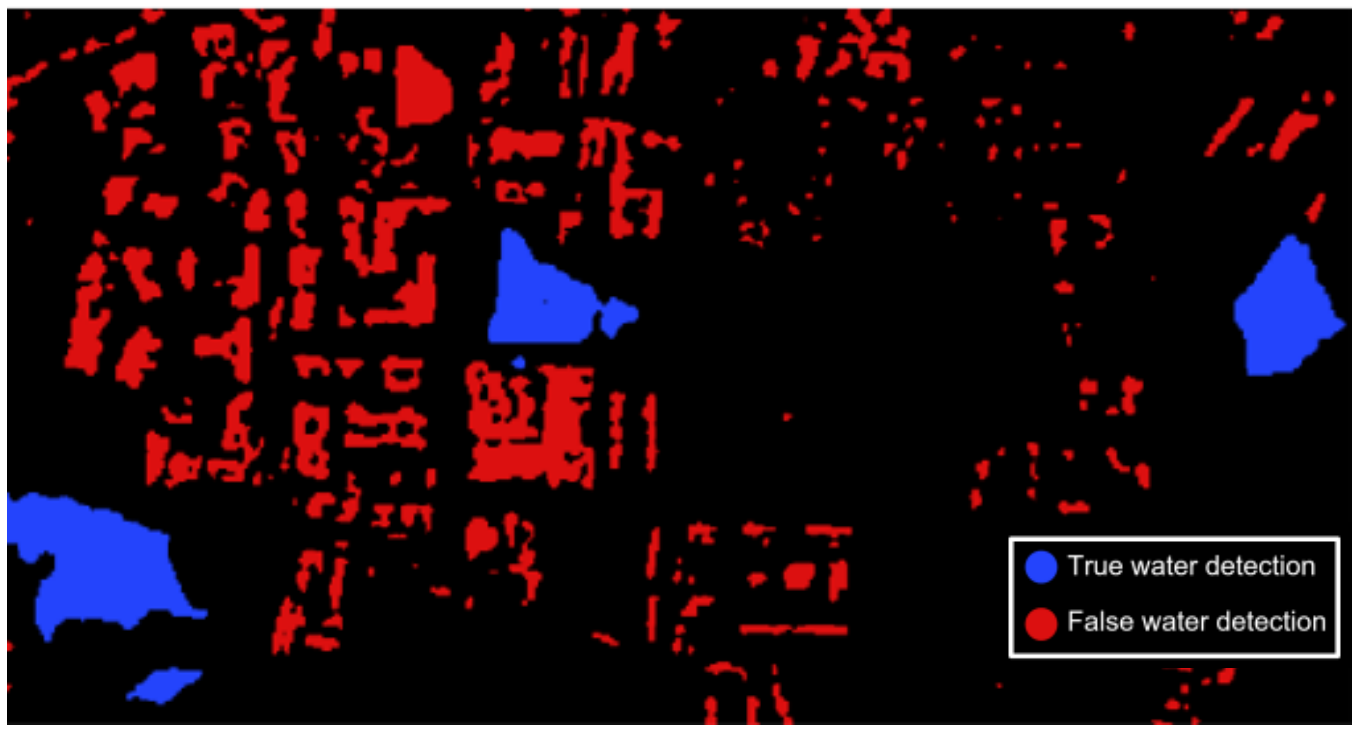

**Fig.7.** Initial annotation based on NDWI

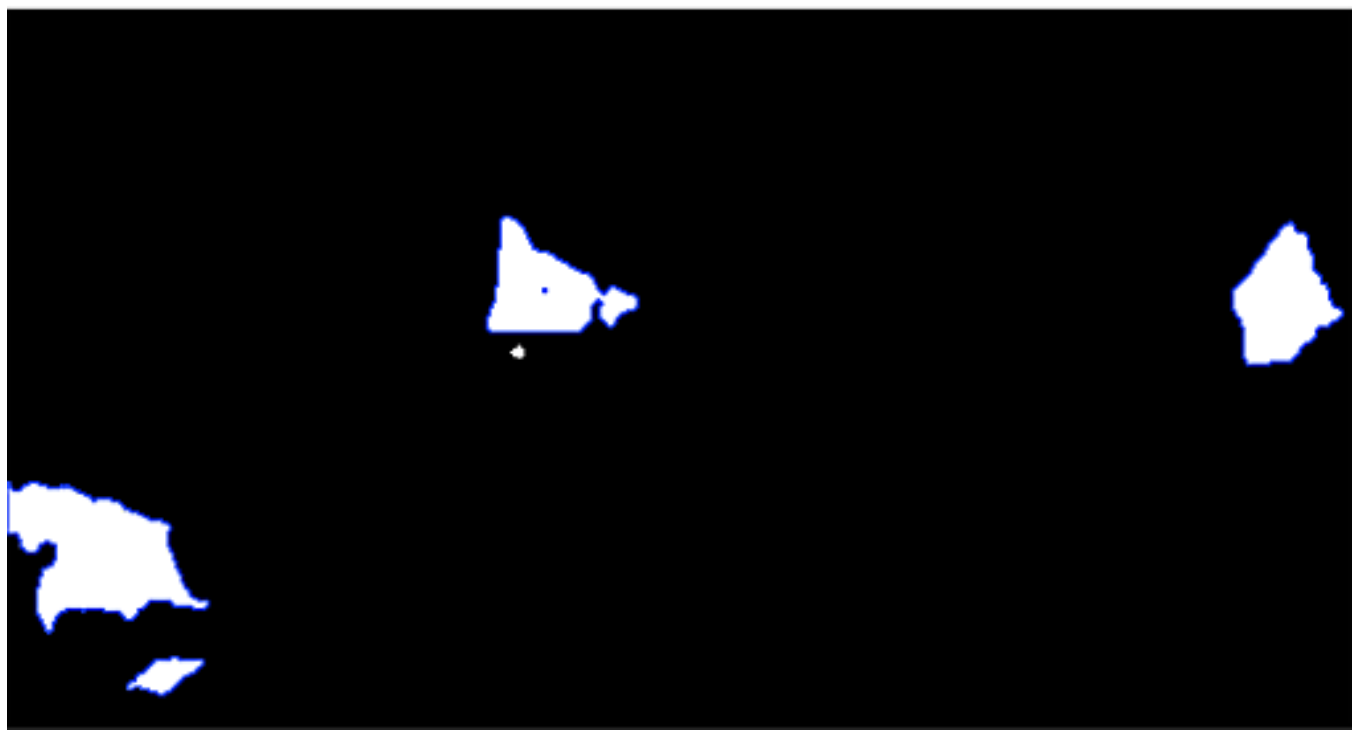

**Fig.8.** Final water body

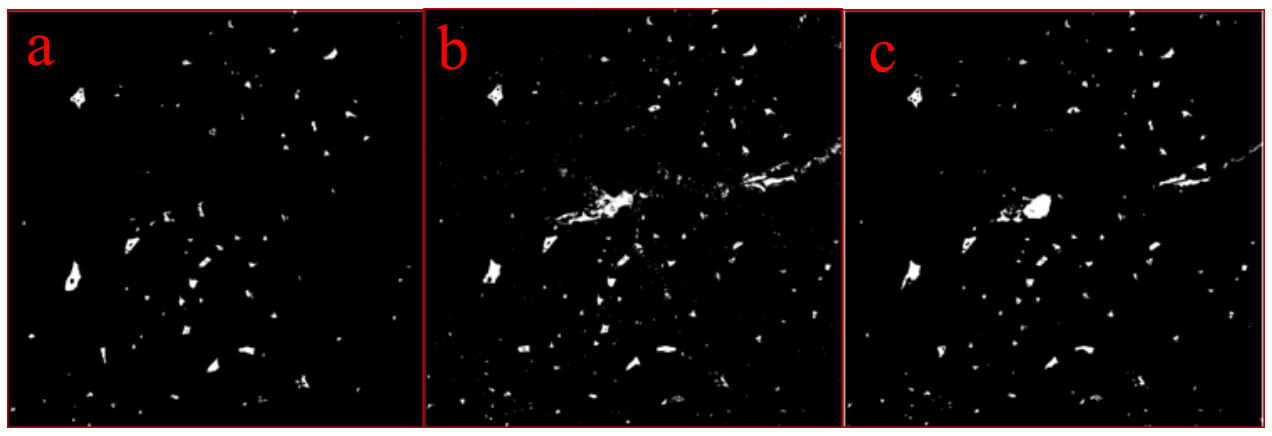

**Fig.9.** Temporal surface water-body. a) before flood b) during flood and c) after flood.

After performing the NDWI band ratio and applying the threshold value, we observed that the identified water body is not accurate as seen in Figure 7. In particularly, we noted that urban areas, especially the pixels representing building tops, were incorrectly classified as water bodies. In order to enhance the precision of the detected water areas, we employed a reclassification technique to eliminate false detections resulting in annotations as seen in Figure 8. This manual correction was carried out using QGIS [16], an open-source software. Special care was taken to annotate areas with known news coverage during the Bengaluru city floods of 2022. Another important point to note was the cloud coverage during these events was high. So the images chosen for annotation were extremely low in cloud cover. This can be addressed in future by performing cloud annotations as well to avoid misclassification. In pre flood, there is 379 hectares area of surface water. it's increased during flood to 776 hectares and again it's drop to 599 hectares. Figure 9 illustrates the changes in annotated water areas before event during event and after event.

## IV. MODEL ARCHITECTURE, OPTIMIZATION, TRAINING AND SELECTION

### *A. Model Architecture and Selection*

The model being used is a Unet [17] architecture with residual connections (ResUNet) [18] for water body segmentation. The model, as shown in Figure 10, has two parts encoder and a decoder. The encoder consists of blocks which down-samples the input using convolution and pooling with residual connections within the blocks. Residual connections are shown to allow training of deeper networks by addressing the problem of vanishing gradients.

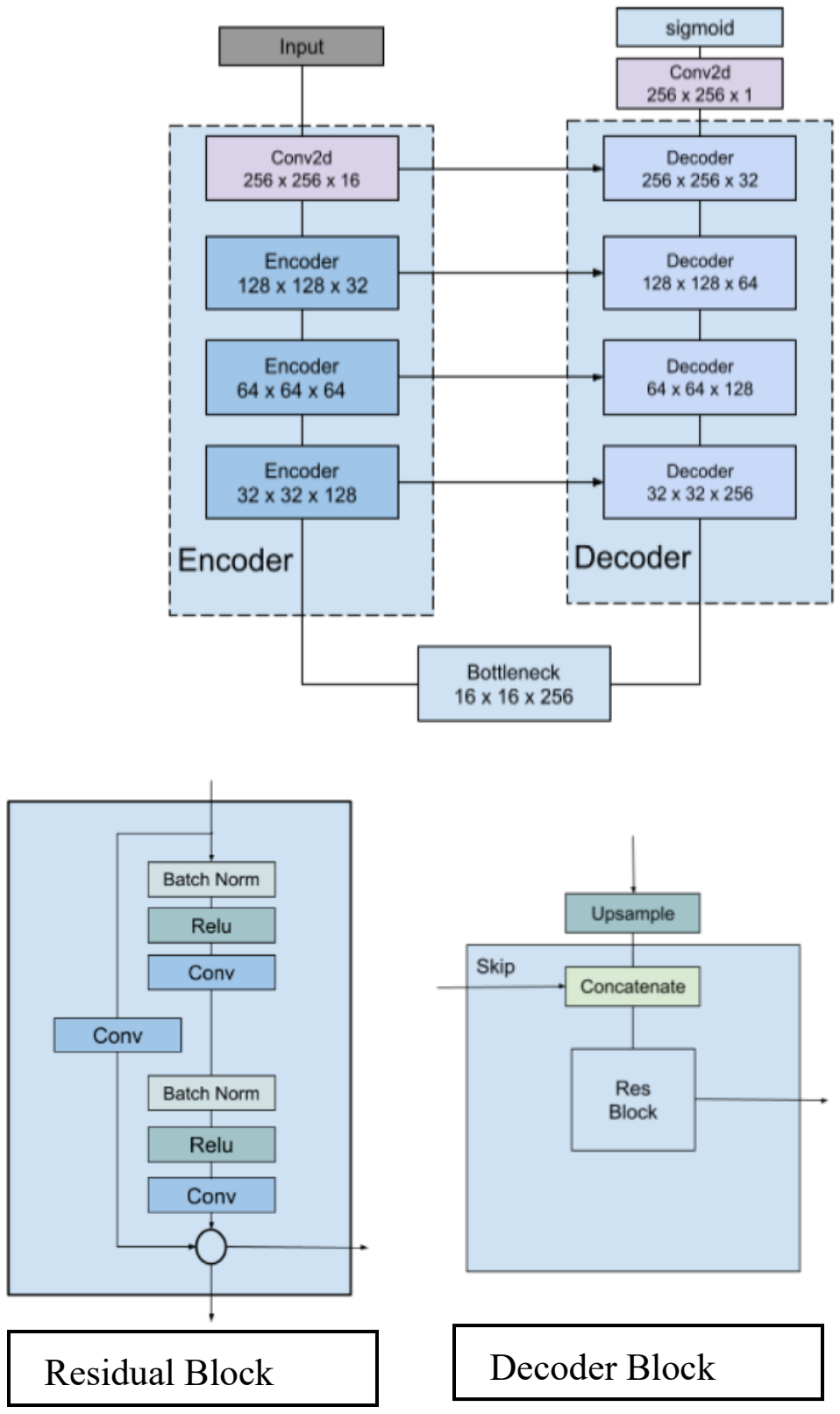

**Fig.10.** ResUNet model architecture diagram.

They are followed by decoder blocks which perform upsampling and concatenation with the skip connection. The skip connections help retain high resolution information from earlier stages . A 1x1 convolution is applied to the final layer to get the output. Sigmoid is applied on the output to get the class values for water body and not water body. The bodies makeup a small portion of the image with the majority of the image containing land area, due to this a model with good binary accuracy does not give good results. Dice coefficient is a metric [19] used in evaluating performance of segmentation tasks, this function is differentiable and can be used as a loss function. Intersection over Union (IoU), Dice metric and binary accuracy were compared and the dice metric was shown to provide better results for water body segmentation. The optimization algorithm Adam was used to train the ResUNet model with learning rate of 0.001. The model was trained for 100 epochs. The batch size of 8 was used during training. To allow accurate assessment of the model results various metrics were used to evaluate the model., the metrics used to evaluate the model were IoU metric, Dice coefficient, Jaccard coefficient and binary accuracy.

### *B. Model Optimization*

Pruning is a technique to reduce a trained large neural network to a smaller network without requiring retraining. Pruning involves removing parameters of a neural network that improves the efficiency of the neural network while trying to maintain the accuracy. This allows deployment in constrained environments such as embedded systems. If pruned networks are retrained it provides the possibility of escaping a previous local minima [20][21] and further improve accuracy. A pruning schedule was initialized using polynomial decay using Tensorflow model optimization library. The pruning schedule parameters chosen are initial sparsity of 0.2, final sparsity of 0.8, begin step of 0 and end step of 5000. Using the Adam optimizer, Dice coefficient Loss as the loss function, the model is compiled and trained for two epochs. Using the pruning schedule the model underwent the pruning process. This allowed for reduction of the initial size of 57.11 MB to the pruned model size of 19.7 MB while preserving accuracy. The pruned model was converted to ONNX (Open Neural Network exchange) [22] to further improve the size to 18 MB while preserving the datatype as float32. A separate process involved a post-training quantization step using TensorFlow's [23] TfLite library from 32 bit precision to 8 bit precision. Post training quantization involves taking a trained model, quantizing the weights, and then re-optimizing the model to generate a quantized model with scales [20]. This reduced the original trained model size from 57.11 MB to 4.8 MB. These steps are summarized in Table I below.

TABLE I
MODEL SIZE OPTIMIZATION IN CONSECUTIVE STEPS

| # | Model | Model Size (MB) |
|---|---|---|
| 1 | Original Trained Model | 57.11 |
| 2 | Pruned Model | 19.7 |
| 3 | ONNX-converted Model | 18 |
| 4 | TfLite quantized model | 4.8 |

### *C. Training Results*

The model achieved the binary accuracy of 0.99 on training and validation and Dice coefficient of 0.93 on training set and 0.83 on the validation set. The overall training and validation metrics are summarized in Table II below.

TABLE II
ORIGINAL MODEL ACCURACY METRICS AFTER 100 EPOCHS

| # | Metric | Training | Validation |
|---|---|---|---|
| 1 | Binary | 0.99 | 0.99 |
| 2 | Dice coefficient | 0.93 | 0.83 |
| 3 | Jaccard Coefficient | 0.88 | 0.70 |

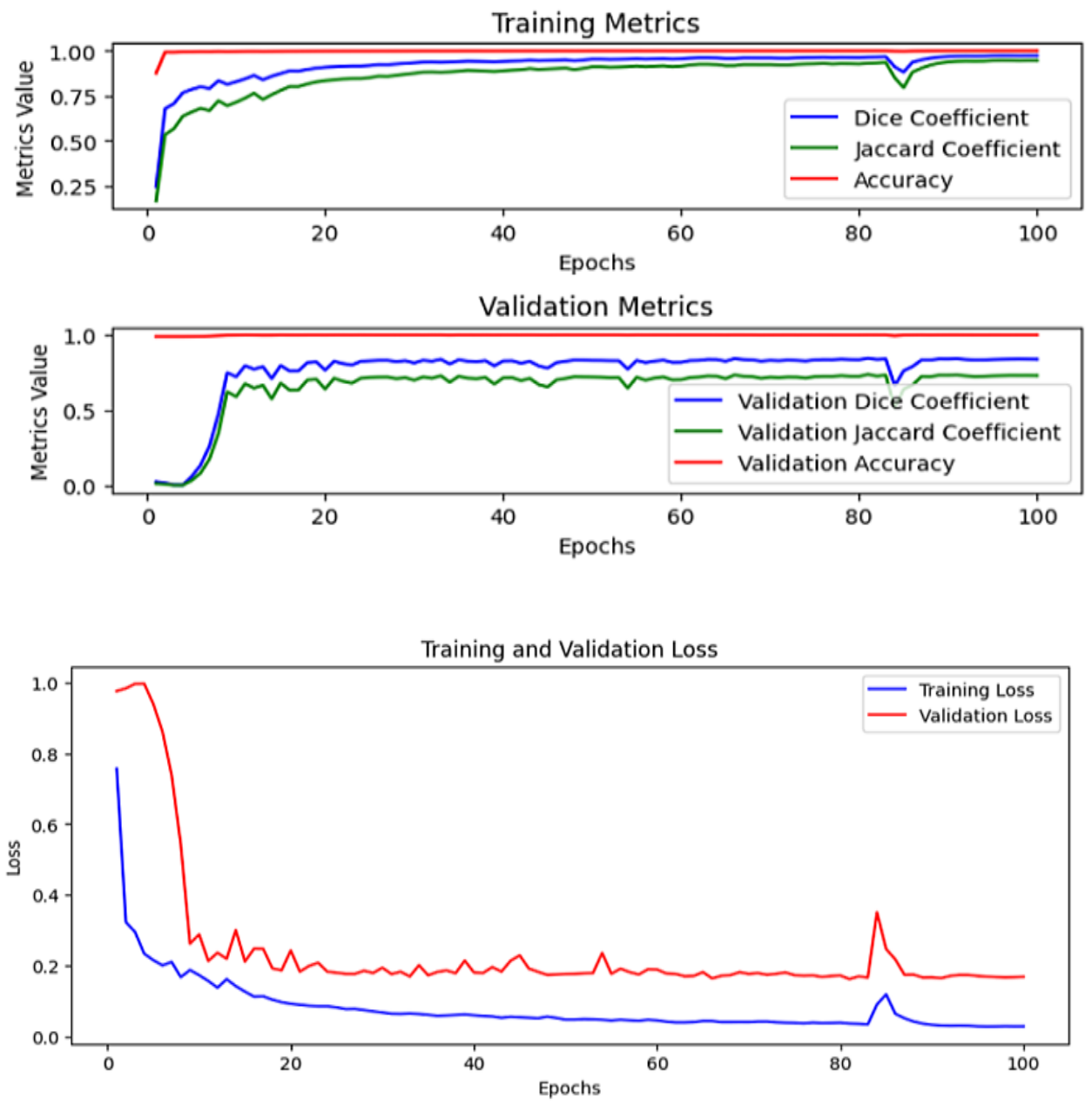

**Fig.11.** Training, Validation metrics and loss values versus epochs.

The training and validation metrics and loss versus the number of epochs can be seen in Figure 11. The validation metric of Dice coefficient of 0.83 was considered acceptable for the purpose of this study.

### *D. Testing on hardware*

It was also important to test the model performance on an edge hardware that can represent the processing capability available on modern satellites onboard. The test run of the model was implemented on an off-the-shelf single board Jetson Tx2i computer from NVIDIA. The board has 8GB 128-bit LPDDR4 and Quad-core Arm® Cortex®-A57 MPCore processor complex.

The choice of CPU (and not GPU) for inferencing was taken because of the lightweight model already performing well with respect to runtimes. The inferencing test was performed for the simulated images, each of size 256 x 256 x 4 in float32 format as mentioned in Section III. For each a run, a batch of 8 images is selected. A test script for running the same batch for 5 times was then executed to get the final runtimes. As documented in Table III below, the average inference time was approximately 49.6 ms. At 4.75m-GSD, a 256 x 256 sized image covers 1216 m x 1216 m spatially or approximately 1.48 sq.km.. In other words, approximately 33 ms per sq.km. for performing detection of flooded areas.

TABLE III

ORIGINAL MODEL RUNTIMES FOR 5 BATCH RUNS

| Run ID | Runtime (s) | Inferencing time per Image (s) |
|---|---|---|
| 1 | 0.3871 | 0.048 |
| 2 | 0.4197 | 0.052 |
| 3 | 0.3940 | 0.049 |
| 4 | 0.4017 | 0.050 |
| 5 | 0.3954 | 0.049 |

## V. ROAD NETWORK INTERSECTION

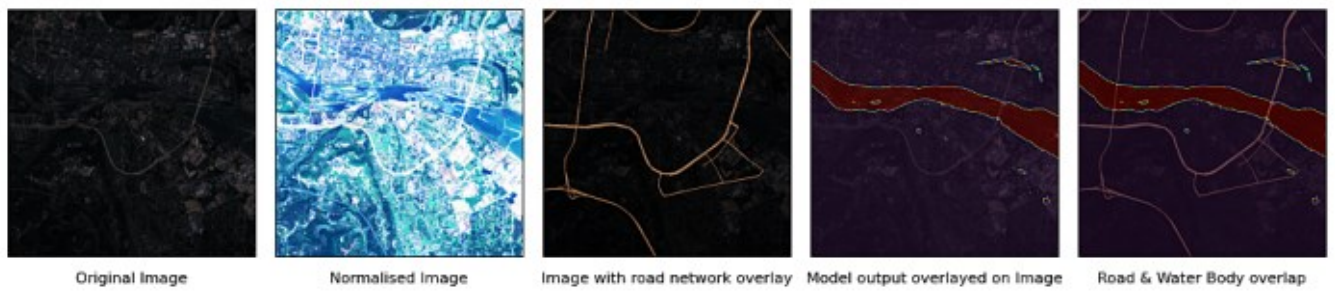

**Fig.12.** Plot for intersection between road networks in an image and water bodies identified by the model.

The onboard computation which generates an image mask with georeferencing data can be considered for processing on an earth-based server where the geometric correction could be first enhanced and an intersection of the water mask areas with the road network can be analyzed. Figure 12 illustrates the same concept. These areas are the ones that might have to be formatted into data which can be made available for navigation services to generate dynamic maps (e.g. GeoJSON format).

Some limitations of the existing method are worth highlighting. Cloud effects can obviously mask certain areas of study and the dataset is best augmented with a cloud mask in the next iteration. It can be challenging to identify flood intersections on underpass roads invisible to the satellite imagery, and on roads that have greater vegetation. Furthermore, there is little attention focused on isolated major roads, and information of these kinds of roads is out of date due to their less frequent traffic.

## V. CONCLUSION

The paper reviewed the application of road health monitoring for navigation services used globally today. The specific case of identification of flooded road segments through onboard computing was discussed. As part of the Orbital AI PhiSat-2 challenge, the process of development of simulated datasets and annotation was described for the city of Bengaluru, India in 2022. A ResUNet model architecture was chosen, trained and optimized for use for edge computing in satellites. The optimized model was testing for processing times on an edge computing machine and the results seem to indicate feasibility for running the model as an onboard application. With the ground-based road network intersection for the flooding event detected onboard, a rapid response on ground can be initiated providing inputs to generate dynamic maps for navigation.

## FUTURE SCOPE OF WORK

It is foreseen that the future work would extend these concepts into a generic road health model (road damage, snow, landslide, etc.) in a constellation of satellites to minimize latency of insights being generated while removing other non-necessary artefacts like clouds etc. Creating the training set especially when dealing with multispectral data can be a challenging process. During the monsoon and cyclone seasons, say, clouds, haze, and noise in satellite images are common. Use of multiple satellite images, high revisit frequencies, and exploration of other spectral bands and sensors (like Synthetic Aperture Radar) to gather real-time data is necessary to address these issues. Several global flooded areas over a period of years are required to be included in the overall dataset emphasizubg higher spatial distribution over frequency. Modern deep learning models, synthetic images, and spectral indices play an important role in obtaining accurate information and validating outcomes.

## ACKNOWLEDGMENT

The authors would like to acknowledge the organizers of the OrbitalAI PhiSat-2 challenge for giving us an opportunity to know more and motivate us about the possibilities unlocked by the PhiSat-2 mission.

**Vishesh Vatsal** is the Chief of Technology of Hyspace Technologies. He received his B.Tech. in Aerospace Engineering from Indian Institute of Technology, Kanpur, India in 2011. He is interested in the application of onboard processing for Earth Observation through multiple satellite missions being executed. The application and software infrastructure for enabling onboard processing is part of SkyServe platform that he is working on.

**Gouranga Nandi,** a Geospatial Engineer at Hyspace Technologies, he received his MSc in Remote Sensing and GIS from Vidyasagar University, Midnapore, India in 2018. He specializes in processing satellite data, constructing models for diverse applications, and actively contributes to EO model development and testing for the SkyServe platform.

**Primo Manilal**, a machine learning intern at Hyspace Technologies, is currently pursuing his Master's degree in computer science. He specializes in computer vision and machine learning.